\documentclass{article}

\usepackage{arxiv}

\usepackage[utf8]{inputenc} 
\usepackage[T1]{fontenc}    
\usepackage{hyperref}       
\usepackage{url}            
\usepackage{booktabs}       
\usepackage{amsfonts}       
\usepackage{nicefrac}       
\usepackage{microtype}      
\usepackage{cleveref}       
\usepackage{lipsum}         
\usepackage{graphicx}
\usepackage{natbib}
\usepackage{doi}

\usepackage[utf8]{inputenc}
\usepackage{graphicx}%
\usepackage{multirow}%
\usepackage{amsthm}%
\usepackage{mathrsfs}%
\usepackage[title]{appendix}%
\usepackage{xcolor}%
\usepackage{textcomp}%
\usepackage{manyfoot}%
\usepackage{booktabs}%
\usepackage{algorithm}%
\usepackage{algorithmicx}%
\usepackage{algpseudocode}%
\usepackage{listings}%

\usepackage{hyperref}%
\usepackage{threeparttable}%

\usepackage{enumitem}
\usepackage{tabularx}
\usepackage{float}
\usepackage{caption}
\usepackage{natbib}
\usepackage{authblk}
\usepackage[justification=raggedright, singlelinecheck=false]{caption}

\newcommand{\customsection}[1]{%
	\subsection*{#1}%
	\vspace{-0.5em}
}

\theoremstyle{thmstyleone}%
\theoremstyle{thmstyletwo}%

\theoremstyle{thmstylethree}%

\title {Beyond Token Limits: Assessing Language Model Performance on Long Text Classification}

\author[1]{Miklós Sebők\thanks{Corresponding author: \texttt{sebok.miklos@tk.hu}}}
\author[1]{Viktor Kovács}
\author[1]{Martin Bánóczy}
\author[2]{Daniel Møller Eriksen}
\author[3]{Nathalie Neptune}
\author[4,3]{Philippe Roussille}
\affil[1]{HUN-REN Centre for Social Sciences, Budapest, Hungary}
\affil[2]{Aarhus University, Aarhus, Denmark}
\affil[4]{3iL Ingénieurs, 43 rue de Sainte Anne, 87015 Limoges, France}
\affil[3]{Institut de Recherche en Informatique de Toulouse,  Toulouse, France}

\begin{document}
\maketitle
	
	\begin{abstract}The most widely used large language models in the social sciences (such as BERT, and its derivatives, e.g. RoBERTa) have a limitation on the input text length that they can process to produce predictions. This is a particularly pressing issue for some classification tasks, where the aim is to handle long input texts. One such area deals with laws and draft laws (bills), which can have a length of multiple hundred pages and, therefore, are not particularly amenable for processing with models that can only handle e.g. 512 tokens. In this paper, we show results from experiments covering 5 languages with XLM-RoBERTa, Longformer, GPT-3.5, GPT-4 models for the multiclass classification task of the Comparative Agendas Project, which has a codebook of 21 policy topic labels from education to health care. Results show no particular advantage for the Longformer model, pre-trained specifically for the purposes of handling long inputs. The comparison between the GPT variants and the best-performing open model yielded an edge for the latter. An analysis of class-level factors points to the importance of support and substance overlaps between specific categories when it comes to performance on long text inputs.
	\end{abstract}

	
	\maketitle

	\section{Introduction}\label{introduction}
	
	In recent years, large language models (LLMs) have revolutionized the
	field of natural language processing (NLP), demonstrating remarkable
	capabilities across a wide array of tasks, from text generation to
	question answering. As these models evolve, a critical question remains:
	Does the length of the input text still significantly impact their
	performance, particularly in classification tasks?
	
	While advances in model architectures and attention mechanisms have
	improved the ability of LLMs to handle longer sequences, the length of
	the input text can still influence the performance and efficiency of
	these models in various ways. Studies have shown a notable degradation
	in reasoning performance with extended inputs, occurring well before the
	models reach their technical maximum input
	capacity\hspace{0pt}\hspace{0pt}. This decline is not merely due to the
	volume of text but also involves the models' ability to focus on and
	retrieve relevant information from within longer sequences \citep{levy_same_2024}.
	
	Transformer-based LLMs rely on attention mechanisms to process input
	sequences. Traditional attention mechanisms have quadratic time and
	memory complexity, making them inefficient for long sequences. To
	address this, models like Longformer \citep{beltagy2020longformer} and BigBird
	\citep{zaheer_big_2020} have introduced sparse attention mechanisms that
	scale linearly with the input sequence length. These mechanisms allow
	the models to capture long-range dependencies without the prohibitive
	computational costs of full self-attention. For
	instance, the Longformer can process sequences up to 4,096 tokens,
	significantly longer than the 512-token limit of models like BERT
	\citep{beltagy2020longformer}.
	
	Recent developments have seen remarkable advancements in generative
	models like GPT \citep{openai_gpt-4_2024} and Llama \citep{touvron_llama_2023}.
	These models have demonstrated significant capabilities in in-context
	learning, generating classification labels based on examples provided in
	their prompts. Despite their large context windows, these models may
	still fall short without fine-tuning, especially for tasks requiring
	deep domain knowledge or nuanced understanding.
	
	The objective of this study is to compare the performance of various
	multilingual encoder models on the Comparative Agendas Project (CAP)
	classification task involving texts that exceed the limit of
	512 tokens. The CAP task classifies multi-domain texts such as policy
	documents, speeches, and news articles based on a standardized topic
	codebook, aiding in the analysis of policy trends and political agenda
	shifts. Numerous other papers have attempted to automate this task using
	fine-tuned LLMs, but little has been written about the case of long
	documents \citep{liu_roberta_2019, sebok_leveraging_2024}. For this purpose,
	we fine-tune two XLM-RoBERTa models \citep{conneau_unsupervised_2020}, differing in
	parameter count, and a Longformer model pre-trained using an XLM-RoBERTa
	checkpoint. We also evaluate and compare proprietary generative AI
	approaches, including zero-shot and one-shot classification using
	GPT-3.5 and GPT-4, as well as an open-source generative model, Llama 3.
	Our aim is to provide a comprehensive evaluation of how these models
	handle long contexts and identify each approach's potential strengths
	and weaknesses.
	
	We demonstrate the critical importance of selecting the appropriate
	dataset for fine-tuning based on text length to achieve optimal
	performance across long texts. However, our results indicate no need for
	specialized methods like the Longformer for long-text classification;
	the existing models are robust enough to handle such tasks. Our key
	findings are as follows:
	
	\begin{enumerate}
		\def\labelenumi{\arabic{enumi}.}
		\item
		We found that employing a combination of short (<512 tokens) and long ($\geq$512 tokens) texts for fine-tuning results in superior performance on long texts compared to using exclusively short or long texts.
		\item
		Reducing the maximum input text length to 512 tokens does not significantly impact the Longformer's classification performance on this particular task.
		\item
		While the Longformer shows marginal improvement over the base XLM-RoBERTa model, the large XLM-RoBERTa outperforms both.
		\item
		Both open and proprietary generative AI, despite leveraging extensive context lengths, fall short in terms of classification performance.
	\end{enumerate}

	In what follows, we review the literature on the CAP task and then
	examine the literature addressing challenges associated with
	long-context sequences. Next, we present the dataset and model selection
	criteria and provide a detailed description of the conducted
	experiments. Subsequently, we evaluate the performance of both open and
	proprietary generative LLMs. We also examine class-level factors
	influencing model performance, such as substance overlaps and support.
	Lastly, we present our findings and compare different models, offering
	recommendations tailored to specific scenarios.
	
	\hypertarget{literature-review}{\section{Literature review}\label{literature-review}}
	
	\hypertarget{leveraging-llms-in-cap-classification}{%
		\subsection{Leveraging LLMs in CAP
			classification}\label{leveraging-llms-in-cap-classification}}
	
	The Comparative Agendas Project (CAP) classification task represents a
	pivotal component of research in leveraging LLMs for text classification
	within the domain of political science. CAP, an initiative that
	systematically codes and tracks issues in public policy across different
	political systems, provides a rich dataset that categorizes political
	text according to a standardized topic codebook.
	
	Research leveraging the CAP coding scheme to study the content of
	various political documents has been a growth industry for more than
	three decades \citep{baumgartner_agendas_1993, baumgartner2019comparative}.
	For instance, scholars have used CAP to explore the issue content of
	questions to ministers \citep{borghetto_agenda_2018, seeberg_power_2023} and
	executive speeches \citep{jungblut_different_2023}. Moreover, the application of
	CAP extends beyond the parliamentary venue, embracing, for instance,
	news media data \citep{thesen_maml_2024}, public opinion \citep{baumgartner_agendas_1993}, and social media data \citep{eriksen_party_2024, russell_tweeting_2021, sebok_leveraging_2024}.
	
	Increasingly, scholars have leveraged LLMs to classify the policy issue
	content of large datasets. Yet, this branch of the literature is still
	in its infancy.\footnote{Check page number 3626 of \cite{sebok2022real} for a
		more detailed summary of this literature.} Among the few studies that
	have taken up this task, \cite{frantzeskakis_legislative_2023} used a BERT
	model to examine the issue-content of laws in African legislatures.
	Similarly, \cite{sebok2022real} employed BERT to categorize Polish
	laws, and \cite{eriksen_party_2024} applied it to classify tweets by Danish MPs.
	Moreover, studies have started utilizing the XLM-RoBERTa model, which
	was introduced to address some limitations and enhance the performance
	of BERT \citep{liu_roberta_2019}. For instance, \cite{sebok_leveraging_2024} have
	launched and validated the CAP Babel Machine, which, by using
	the XLM-RoBERTa model, has been shown to produce state-of-the-art
	outputs for domains like news media or speeches in parliament.
	
	However, these studies share a critical limitation, that is, the
	reliance on models that can only process up to 512 tokens. This
	constraint significantly hampers the analysis of longer documents, such
	as laws or draft legislation, which can span several hundred pages.
	Addressing this challenge by validating models capable of processing
	longer documents is crucial.
	
	\hypertarget{challenges-of-using-long-context}{%
		\subsection{Challenges of using long context}\label{challenges-of-using-long-context}}
	
	Processing long input sequences in LLMs presents significant challenges.
	The core issue is the self-attention mechanism of Transformer blocks,
	which, while enabling interactions among input elements, incurs a
	quadratic computational cost with respect to the sequence length. This
	high computational demand presents a complexity challenge for language
	models, which necessitates advanced computational resources and
	sophisticated algorithmic strategies to manage efficiently. Furthermore,
	the requirement for complex reasoning to sift through and synthesize
	relevant information from extended contexts further complicates the use
	of long input sequences in practical applications.
	
	To address these computational and operational challenges, several
	strategies have been developed. Recurrence and memory compression
	techniques have been proposed as effective means to optimize memory
	usage and reduce the computational overhead associated with
	self-attention \citep{chen_walking_2023, tang_survey_2024}. Additionally, the
	introduction of sparse attention patterns, as exemplified in models like
	Longformer \citep{beltagy2020longformer} and GPT-3 \citep{brown_language_2020},
	significantly cuts down the number of necessary interactions between
	input elements, thereby enhancing processing efficiency. Moreover, the
	development of kernels and approximate methods \citep{choromanski_rethinking_2022, katharopoulos_transformers_2020} for self-attention offers viable,
	computationally less intensive alternatives to the traditional,
	resource-heavy self-attention mechanism.
	
	Presently, large language models are capable of processing long input
	sequences, with the ability to handle tens of thousands of tokens.
	Claude 3.5 Sonnet, for example, has a 200K token context
	window\footnote{\url{https://www.anthropic.com/news/claude-3-5-sonnet}}.
	However, the performance of these models on tasks requiring the
	processing of very long input sequences varies significantly based on
	the position of the relevant information within the input context \citep{, liu_lost_2024}. Limited work has been done in evaluating the
	performance of large language models on very long input sequences,
	indicating a gap in this area.
	
	\hypertarget{understanding-encoder-and-decoder-architectures}{%
		\subsection{Understanding encoder and decoder architectures}\label{understanding-encoder-and-decoder-architectures}}
	
	Decoder LLMs, like those from the GPT and Llama series, are capable of
	processing long contexts due to their autoregressive architecture, which
	allows for efficient scaling of the context window size. Unlike encoder
	models, which are typically limited to fixed sequence lengths, decoder
	models predict the next token in a sequence by leveraging previously
	generated tokens as context. This autoregressive nature simplifies the
	training objective, making it easier for models to handle and extend to
	longer context windows (e.g., GPT-4's ability to handle 128,000 tokens).
	The self-attention mechanism in decoder models is designed to prioritize
	relevant tokens from earlier parts of the sequence, which allows for
	improved performance with long-range dependencies, even though models
	sometimes struggle with mid-context information retrieval \citep{liu_lost_2024}.
	
	In contrast, encoder models employ bidirectional encoder architecture,
	allowing them to consider the entire context (both left and right)
	simultaneously when predicting masked tokens. This bidirectional
	approach, coupled with its masked language model (MLM) training
	objective, enables them to capture richer contextual information, making
	it highly effective for text classification tasks. Additionally, the
	pre-training of BERT-based models includes a next sentence prediction
	(NSP) task, enhancing its understanding of sentence relationships, which
	further benefits tasks that require deep comprehension of text, such as
	classification. Fine-tuning these models involves passing the contextual
	representations through a simple feed-forward neural network, often
	leading to superior performance in classification tasks compared to GPT
	due to its comprehensive context modelling \citep{gasparetto_survey_2022}.
	However, they have a fixed input size, limited to 512 tokens. This limit
	is technically 510 tokens of user-provided text plus 2 special tokens
	(usually {[}CLS{]} at the beginning and {[}SEP{]} at the end). This
	limitation means that BERT-based models can struggle with tasks
	requiring comprehension of longer texts due to the truncation at the
	maximum sequence length.
	
	\hypertarget{an-overview-of-encoder-models-for-long-sequences}{%
		\subsection{An overview of encoder models for long sequences}\label{an-overview-of-encoder-models-for-long-sequences}}
	
	In recent years, several new encoder architectures have been developed
	to address the challenge of long input sequences. The
	\textbf{Longformer}, developed by \citep{beltagy2020longformer}, is particularly
	noteworthy for its modification of the self-attention mechanism to scale
	linearly, enabling the processing of documents with thousands of tokens.
	It uses a hybrid approach combining local windowed attention and
	task-motivated global attention. The local attention operates within
	fixed windows around each token, while global attention allows specific
	tokens to attend to the entire sequence. This efficient combination
	allows the Longformer to handle long documents without losing contextual
	information due to partitioning or truncation. The model has
	outperformed other transformer models like RoBERTa on long-document
	tasks and sets benchmarks on datasets such as WikiHop and TriviaQA.
	
	The Longformer-Encoder-Decoder (LED) further extends this capability to
	sequence-to-sequence tasks like summarization \citep{beltagy2020longformer}.
	Other models have also introduced innovative solutions for handling long
	sequences. \textbf{Big Bird} uses a sparse attention mechanism, reducing
	the quadratic complexity of traditional transformers to linear by
	incorporating global, local, and random attention patterns. This allows
	it to handle sequences significantly longer than previous models while
	maintaining strong performance on tasks like summarization and question
	answering \citep{zaheer_big_2020}. \textbf{MEGA} adopts a moving average
	and gating mechanism to manage long-range dependencies efficiently,
	reducing computational demands while retaining contextual information
	\citep{ma_mega_2023}.
	
	Similarly, \textbf{Hierarchical Transformers} process long documents by
	breaking them into smaller chunks, which are fed into a recurrent or
	transformer layer, effectively preserving key dependencies \citep{pappagari_hierarchical_2019}. \textbf{XLNet} builds on Transformer-XL by using a dilated
	attention mechanism and permutation-based training to process sequences
	of up to one million tokens, making it suitable for tasks requiring
	extremely long-context modelling \citep{yang_xlnet_2020}. These models,
	though varied in approach, all aim to improve the efficiency and
	scalability of long-context processing in transformers. However, their
	practical usability is currently limited. As shown in Table 1, at the
	time of writing, only Longformer offers a multilingual pre-trained model checkpoint.
	
	
	\begin{threeparttable}[H]
		\scriptsize
        \caption{Encoder architectures addressing the issue of long contexts}
		\begin{tabularx}{\textwidth}{@{} 
				>{\hsize=0.8\hsize}X  
				>{\hsize=0.7\hsize}X
				>{\hsize=0.4\hsize}X
				>{\hsize=1.3\hsize}X
				>{\hsize=1.0\hsize}X
				>{\hsize=0.5\hsize}X
				>{\hsize=0.8\hsize}X
				@{}}  
			\toprule
			\textbf{Model} & \textbf{Author} & \textbf{Year} & \textbf{Key Technique} & \textbf{Advantages} & \textbf{Max Length} & \textbf{Multilingual} \\
			\midrule
			Longformer & Beltagy et al. & 2020 & Sparse attention, combining local and global attention & Linear complexity & 4096 & Yes \\
			BigBird & Zaheer et al. & 2020 & Sparse attention & Linear complexity & 4096 & No \\
			Moving \newline Average \newline Gated  \newline Attention & Ma et al. & 2023 & Moving average equipped gated attention & Linear complexity & 4096 & No \\
			Hierarchical Transformers & Schamrje\newline et al. & 2021 & Hierarchical attention & Improved contextual understanding & 4096 & No \\
			Transformer-XL (XLNet) & Dai et al. (Yang et al.) & 2019 (2020) & Segment-level recurrence & No theoretical max length & - & No \\
			\bottomrule
		\end{tabularx}
		
	\end{threeparttable}
	
	\hypertarget{data-and-methods}{%
		\section{Data and Methods}\label{data-and-methods}}
	
	\hypertarget{data-selection}{%
		\subsection{Data Selection}\label{data-selection}}
	
	We downloaded publicly accessible data from the website of the
	Comparative Agendas Project to assemble a multilingual and multidomain
	dataset to fully leverage the transfer learning capabilities of the
	language models used. This dataset, which we will refer to as the
	"pooled dataset," primarily consists of Hungarian documents, with
	additional English and Dutch documents and smaller
	proportions of French and Italian documents. The majority of documents
	were categorized into the domains of parliamentary speeches, media
	reports, or legislative texts, with smaller representations of executive
	speeches and executive orders. In assembling the dataset, we aimed to
	include texts of varying lengths, ranging from single sentences (e.g.,
	newspaper titles) to complete documents (e.g., legislative texts).
	
	The texts in the dataset are hand-coded using 21 Comparative Agendas
	Project (CAP) labels. Note that we removed rows containing no policy
	content during our experiments. We also excluded the label 'Culture'
	(23) during sampling as some of the source datasets did not feature this
	label. The labels within the dataset exhibited a highly imbalanced
	distribution, a common phenomenon with CAP-coded corpora. Text
	preprocessing was minimal, as the large language models used are robust
	to noise in the fine-tuning data. We removed trailing whitespaces,
	additional tabulations, and newline characters. Texts containing fewer
	than five words were dropped, as we considered these examples too short
	to provide relevant information.
	
	For model fine-tuning, we generated three distinct samples from the
	pooled dataset, categorized by text length. To ensure comparability, the
	sampling procedure was designed to create fine-tuning datasets of equal
	size (150,661 rows) and with nearly identical proportions across various
	language-domain combinations (Figure 1). We quantified text length by
	annotating each text with the number of tokens using the XLM-RoBERTa
	tokenizer. The resulting datasets were categorized as follows: the
	`short' dataset, comprising texts with fewer than 512 tokens; the `long'
	dataset, consisting of texts with 512 or more tokens; and the `mixed'
	dataset, which includes an equal proportion of texts from both the
	`short' and `long' datasets. The token count distribution for each
	fine-tuning dataset is shown in Figure 2. The test data consisted solely
	of long documents, totalling 40,000 examples, with the language, domain,
	and label distribution matching that of the fine-tuning data.
	
	\begin{figure}[H]
		\centering
		\includegraphics[width=300pt]{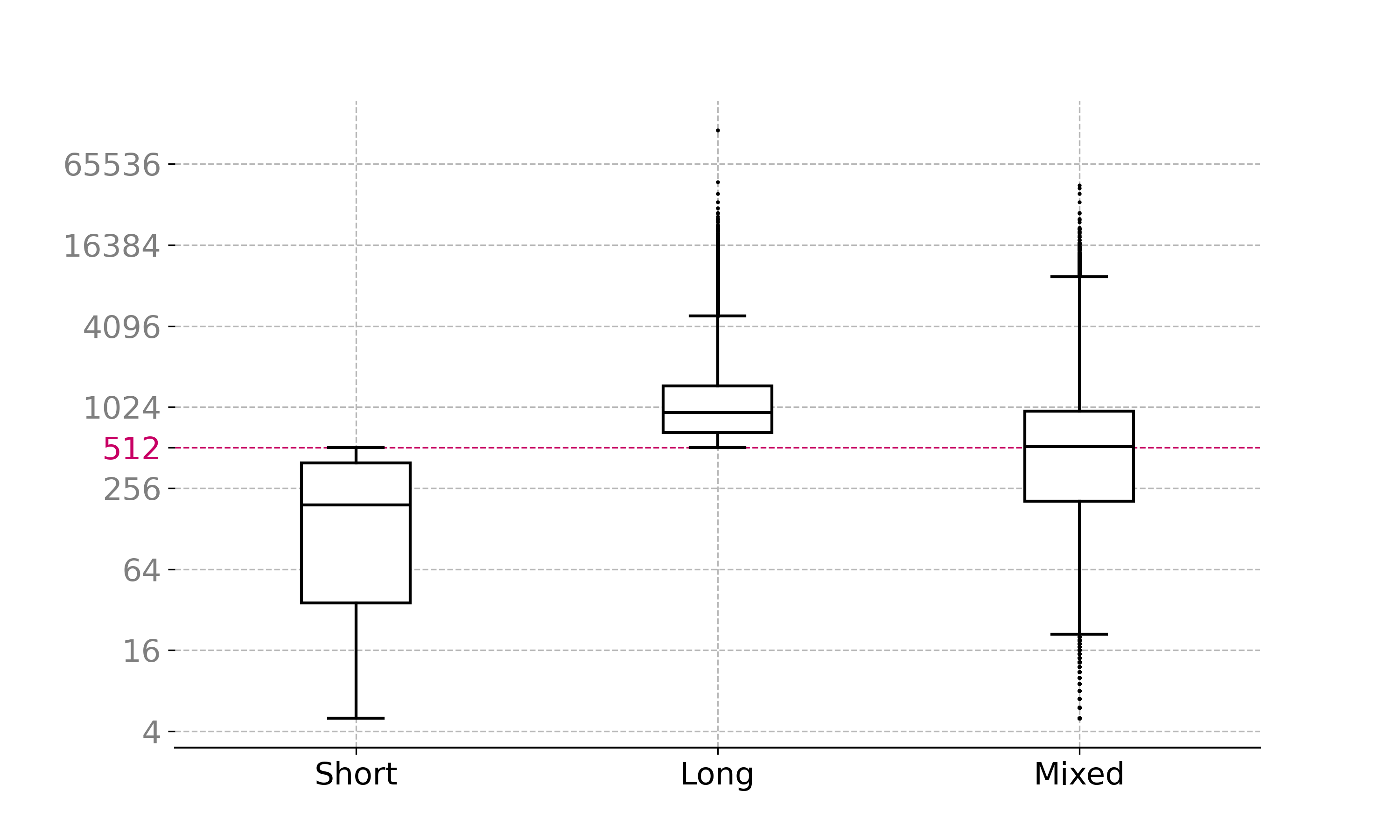}
		\caption{Token count distribution of the fine-tuning datasets\protect\footnotemark}
	\end{figure}
	
	\footnotetext{We used a log2 scale to visualize the token count distribution, as the long dataset contains outliers that would otherwise distort the figure.}
	
	\begin{figure}[H]
		\centering
		\includegraphics[width=280pt]{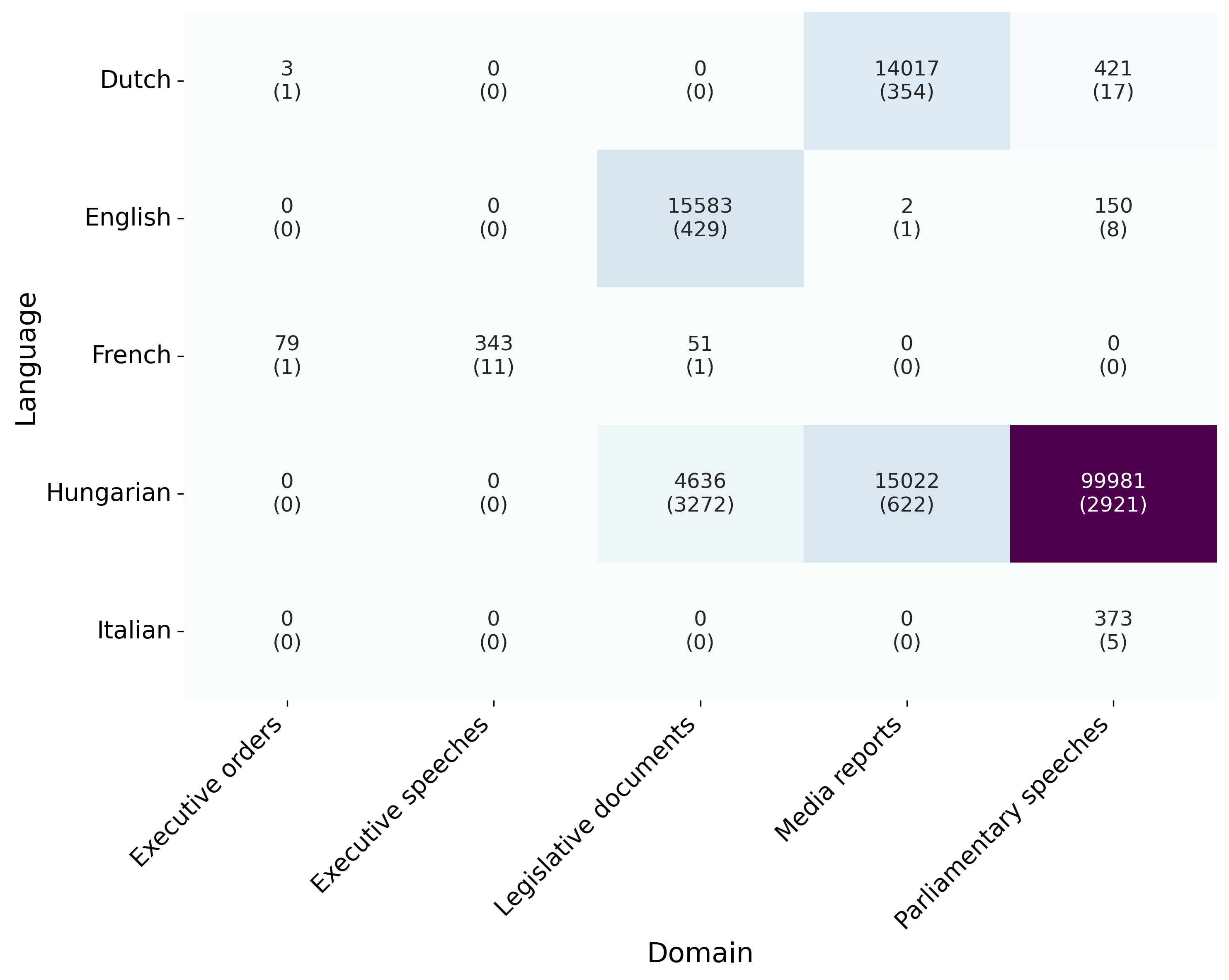}
		\caption{Examples per language-domain combinations in fine-tuning datasets\protect\footnotemark}
	\end{figure}
	
	\footnotetext{The heatmap displays the average number of documents, with the standard deviation in parentheses, for each language-domain combination across the three fine-tuning datasets.}

	\hypertarget{model-selection}{%
		\subsection{Model selection}\label{model-selection}}
	
	We selected the Longformer model to examine whether it improves the
	performance of policy topic classification for long texts compared to
	standard encoder models. The Longformer's maximum sequence length makes
	it well-suited for our task, as the majority of the texts we observed
	contain fewer than 4096 tokens. Several other models were considered but
	found unsuitable due to language limitations. Big Bird, although it can
	also handle contexts up to 4096 tokens, is fine-tuned with the same data
	as RoBERTa which means it's a monolingual English model, thus lacking
	the multilingual capabilities required for our comparison. A similar
	issue arises with XLNet and models utilizing Hierarchical Attention and
	MEGA. These constraints restrict these models' applicability in a
	setting where understanding and processing multiple languages is
	crucial. Furthermore, the sliding window technique, which could be used
	to implement XLM-RoBERTa models, was not considered because previous
	results indicated a decrease in performance compared to baseline models
	(Mate et al., 2023). We also evaluated and compared proprietary
	generative models, using zero-shot and one-shot classification using
	GPT-3.5 Turbo and GPT-4 Turbo, as well as an open-source generative
	model, Llama 3. We used a stratified sample of 500 examples to evaluate
	these generative models due to cost and resource constraints.
	
	\hypertarget{fine-tuning-and-evaluation}{%
		\subsection{Fine-tuning and
			evaluation}\label{fine-tuning-and-evaluation}}
	
	We employed a similar approach for fine-tuning both the XLM-RoBERTa and
	Longformer models. We set the default number of epochs at 10 and
	implemented early stopping, terminating the fine-tuning process after
	two consecutive epochs without improvement in validation loss. The
	validation data was sampled from the test data\footnote{This approach
		was adopted due to time and resource constraints, and the validation
		data was used exclusively for early stopping. While this method can
		provide a quick indication of model performance, it may lead to an
		optimistic estimate of the model's effectiveness on unseen data.
		Future work should address this limitation to provide a more
		comprehensive assessment of model performance.}. Minor hyperparameter
	changes were not significantly influencing model performance but were
	important due to certain limitations of the hardware that we were using.
	Batch sizes were chosen based on the resource needs of the respective
	models; for XLM-RoBERTa, we used a batch size of 32, while for
	Longformer, we used a batch size of 8. For the fine-tuning process, we
	utilized NVIDIA A40 and V100 GPUs, and in the case of the Longformer,
	A100 GPUs were used. Depending on the experiment, truncation was applied
	during the tokenization of texts, with maximum lengths set at 512, 1024,
	and 2048.
	
	To address data imbalances, we selected the weighted macro F1-score as
	our main evaluation metric. This metric effectively combines precision
	and recall into a single measure, computed separately for each class
	before being averaged, with the contributions of each class weighted by
	the number of examples. This method ensures equitable contributions from
	all classes to the overall metric, regardless of their frequency,
	thereby preventing dominant classes from skewing the results.
	
	We conducted four experiments to address our research question related to the effect of input text length on model performance (please refer to the Appendix for more details on the experiments).
	
	\subsection{Experiment overview}\label{experiment-overview}
	
	\customsection{Experiment 1: Determining optimal training text length composition}
	
	In determining the optimal dataset composition concerning text length
	distribution, we assessed three variants of the xlm-roberta-base model;
	each fine-tuned using pre-sampled datasets of short, long, and mixed
	texts. Our objective was twofold: 1) to determine whether exclusive use
	of long texts enhances model performance in classifying such texts, and
	2) to investigate how a combination of short and long texts improves or
	degrades model performance.
	
	\customsection{Experiment 2: Effect of maximum input sequence length}
	
	To investigate how the maximum input sequence length affects the model's
	classification of long documents, we fine-tuned and evaluated three
	Longformer (markussagen/xlm-roberta-longformer-base-4096) models with
	truncation at 512, 1024, and 2048 tokens. We excluded 4096 tokens
	because the majority of documents contain fewer than 2048 tokens. Models
	were fine-tuned using the long dataset, for it exclusively includes
	texts that are 512 tokens or longer.
	
	\customsection{Experiment 3: Comparison of model architectures and sizes}
	
	To determine if the Longformer model
	(markussagen/xlm-roberta-longformer-base-4096), with its longer maximum
	sequence length, improves the classification of long documents compared
	to XLM-RoBERTa (xlm-roberta-base), we fine-tuned both models using the
	mixed dataset. This choice was informed by the superior performance
	observed in Experiment 1 on long documents. For XLM-RoBERTa, the
	truncation length was set to the default 512 tokens, whereas for the
	Longformer, we utilized 2048 tokens. Compared to the first experiment,
	which focused on determining the optimal dataset composition concerning
	text length distribution using XLM-RoBERTa, this third experiment
	specifically compares the capabilities of XLM-RoBERTa and Longformer in
	handling long documents.
	
	It is crucial to note that in our previous experiments, we compared
	models of equal size. The question remains: How does an increased
	parameter count influence the classification of long texts? To address
	this within Experiment 3, we employed two versions of XLM-RoBERTa with different parameter
	counts: base (125M) and large (355M) for comparison, both fine-tuned
	using the mixed dataset.
	
	\customsection{Experiment 4: Comparison with generative pre-trained models}
	
	Proprietary models offer larger context windows and impressive benchmark
	performance, though they come with the disadvantage of being black boxes
	and yielding outputs that are not fully deterministic. Additionally, we
	lack detailed information regarding the precise language coverage of the
	data used for the pre-training of these models. However, it is confirmed that
	the majority of the data originates from English language sources and
	exhibits worse performance on low-resource languages (Robinson et.al,
	2023). In terms of context, GPT-3.5-Turbo has different versions, one
	with a context window of 4096 tokens and one with 16,384 tokens, while
	GPT-4 features a notable 128,000 tokens.
	
	In Experiment 4, we aimed to assess whether the expansive context windows of generative
	LLMs offer improved classification performance for long texts compared
	to earlier models. Due to time, resource, and cost constraints, we had
	to undersample the long test set from our previous experiment. We
	generated two stratified samples of 500 examples each for the English and
	Hungarian languages. This approach allowed us to highlight
	classification performance differences between high-resource and
	low-resource languages while evaluating models. As a benchmark, we chose
	xlm-roberta-base fine-tuned using mixed data. Additionally, we evaluated
	markussagen/xlm-roberta-longformer-base-4096 (mixed). For generative
	models, we selected three models: GPT-3.5 Turbo and GPT-4 Turbo
	(accessible via the OpenAI API) and an instruction-tuned version of
	Llama 3\footnote{The specific models we used were
		\emph{gpt-3.5-turbo-0125}, \emph{gpt-4-turbo-2024-04-09} and
		\emph{Llama-3-8B-Instruct}. To ensure the model outputs are as
		deterministic as possible, we set a fixed seed and a temperature of 0.}.
	
	Two classification methods were employed: zero-shot classification,
	where only labels and the text to be classified were provided to the
	model, and one-shot classification, where one example per label was also
	provided to the model. For the OpenAI models, each classification
	example was provided as a separate request. GPT-3.5 Turbo and Llama 3
	were used for zero-shot classification; the former did not support a
	context window large enough to include examples (importantly, these are
	long texts) for each label, while the latter required memory exceeding
	our current hardware capabilities. Zero-shot classification was
	conducted exclusively using English data with GPT-4 Turbo.
	
	\hypertarget{results}{%
		\section{Results}\label{results}}
	
	This section presents the findings from our experiments on long-text
	classification, focusing on the impact of fine-tuning data, truncation
	length, and model architecture. We explore how different text lengths
	and model sizes affect performance, using weighted macro F1 as the
	primary metric, with precision and recall for additional insights. The
	experiments also compare the performance of generative models like GPT
	and Llama against BERT-based models, such as XLM-RoBERTa and Longformer,
	to determine optimal strategies for fine-tuning models on long documents
	across various languages.
	
	\hypertarget{text-length-distribution}{%
		\subsection{Text length
			distribution}\label{text-length-distribution}}
	
	The first experiment focused on the impact of different text length
	distributions ---short (\textless512 tokens), long (\textgreater=512
	tokens), and mixed--- in the fine-tuning data on model performance when
	classifying long documents. As Figure 3 shows, the relatively poor
	baseline performance (0.69 F1) of the xlm-roberta-base model fine-tuned
	using short texts improves significantly (0.75 F1) when we swap the
	short fine-tuning dataset to the long one. Using a mixture of both text
	lengths also results in minor improvement (0.76 F1) compared to solely
	long texts. 
	
	Taking a look at other metrics, it's important to note that
	precision and recall are nearly identical for the long and mixed cases
	and show only a slight difference in the case of the model fine-tuned
	using the short dataset, resulting in a balanced performance with no
	tradeoff between metrics. These results suggest that while in most cases
	using solely long documents to fine-tune the model already results in
	significant improvement in metrics, using a mixture of both short and
	long texts might be the best way to fine-tune a model for long text
	classification.
	
	\begin{figure}[H]
		\centering
		\includegraphics[width=320pt]{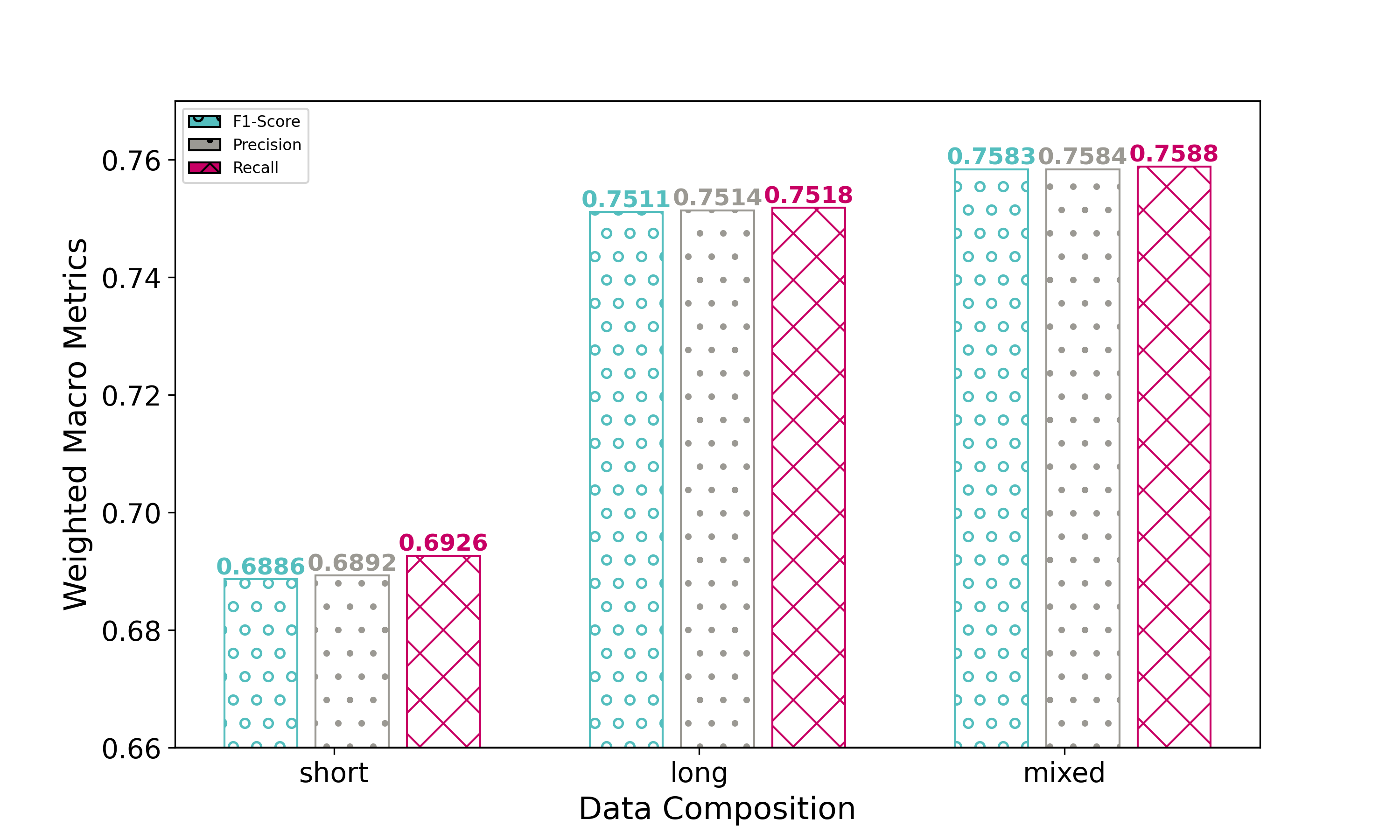}
		\caption{Models fine-tuned using different data compositions.}
	\end{figure}
	
	\subsection{Effect of input sequence
		length}
	
	To explore the impact of maximum input sequence length on the classification of long documents, we fine-tuned and evaluated three
	Longformer models using long data both for fine-tuning and evaluation.
	Truncation was set at 512, 1024, and 2048 tokens. As Figure 4 shows,
	choosing the maximum length did not significantly affect performance.
	This could be explained by the fact that approximately 83\% of the data
	consisted of texts with fewer than 2048 tokens. The results imply that
	the policy topic of a text can often be identified within the first 512
	tokens in these cases.
	
	\begin{figure}[H]
		\centering
		\includegraphics[width=320pt]{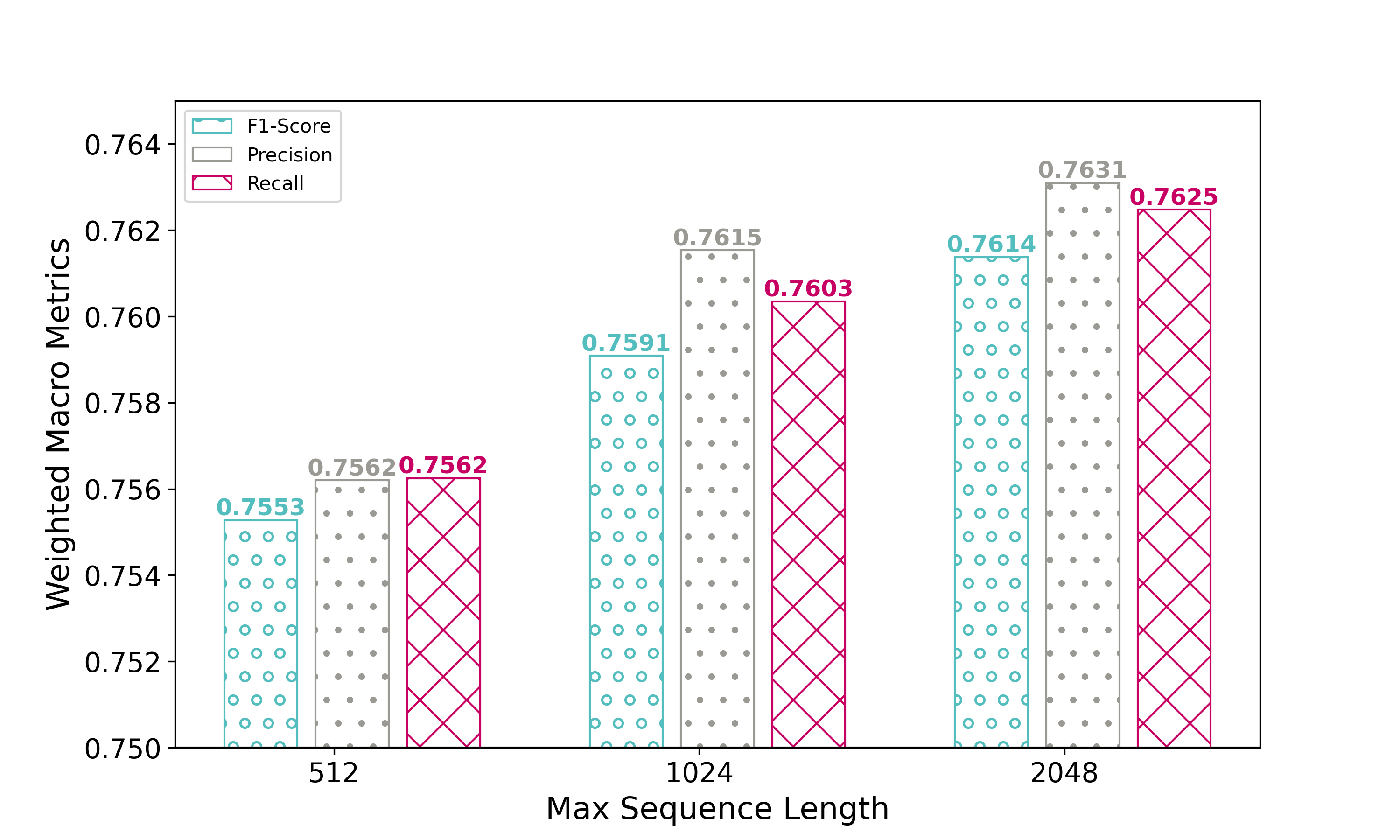}
		\caption{Models fine-tuned using different truncation options.}
	\end{figure}

	\hypertarget{model-architectures-and-sizes}{%
		\subsection{Model architectures and
			sizes}\label{model-architectures-and-sizes}}
	
	To assess whether the Longformer model
	(\emph{markussagen/xlm-roberta-longformer-base-4096} with a truncation
	length of 2048) enhances the classification of lengthy documents
	compared to XLM-RoBERTa (base), we fine-tuned both models using a mixed
	dataset. Additionally, we fine-tuned an XLM-RoBERTa (large) model to
	evaluate the impact of parameter count on performance. Figure 5
	illustrates that xlm-roberta-large outperformed all other models, likely
	due to its increased parameter count. Our results align with those of
	\cite{kaplan_scaling_2020}, which suggest that larger models generally perform
	better. Figure 6 presents label-level F1-scores, indicating varying
	rates of improvement between the Longformer model, xlm-roberta-base, and
	xlm-roberta-large. These results warrant additional experiments to tease out label level effects. We take up this task in the Discussion below.
	
	\begin{figure}[H]
		\centering
		\includegraphics[width=320pt]{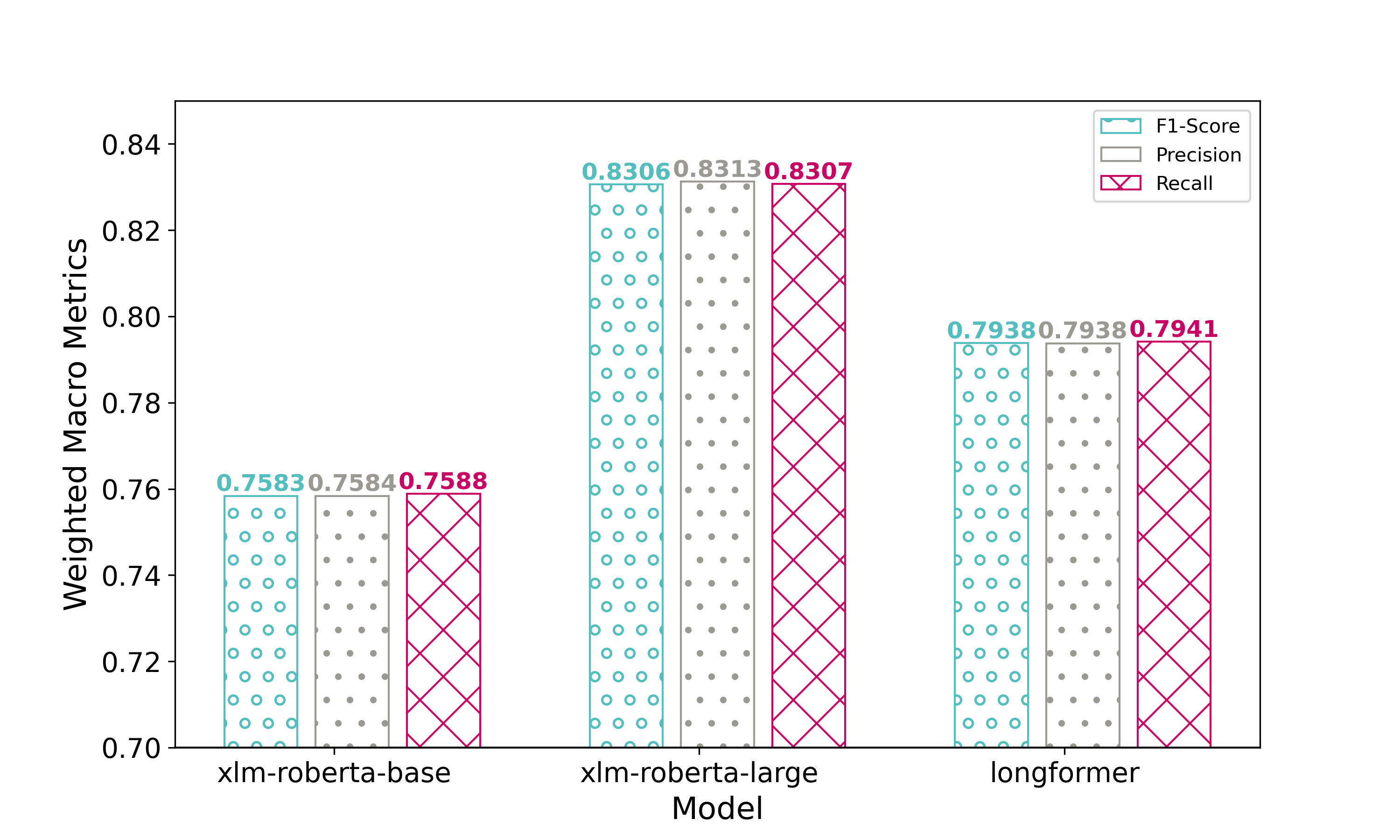}
		\captionof{figure}{Comparison of the Longformer and the XLM-RoBERTa variants.}
	\end{figure}
	
	\begin{figure}[H]
		\includegraphics[width=340pt]{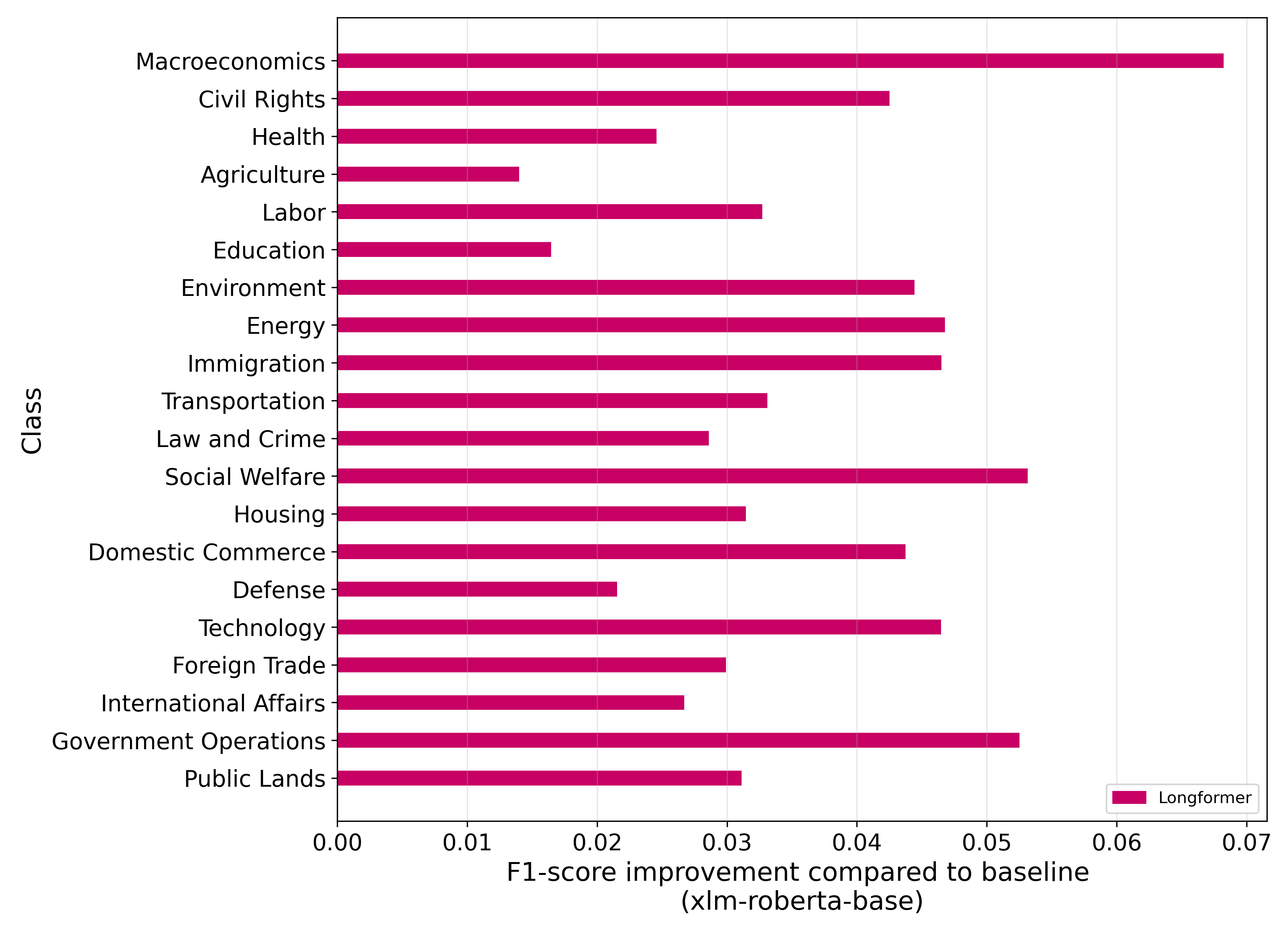}
		\caption{Changes in label-level F1-score}
	\end{figure}
	
	\hypertarget{generative-models}{%
		\subsection{Generative models}\label{generative-models}}
	
	Finally, we also aimed to evaluate if the extended context windows of generative LLMs
	enhance classification performance for long texts compared to BERT-based
	models. Due to constraints in time, resources, and costs, we used two
	undersampled versions of the long test set. We created two stratified
	samples of 500 examples each for English and Hungarian languages. This
	method enabled us to assess performance disparities between
	high-resource and low-resource languages while comparing multiple
	models.
	
	\begin{threeparttable}[H]
		\centering
		\scriptsize
        \caption{Weighted macro F1 comparison of BERT-based models and generative models}
		\begin{tabularx}{\textwidth}{@{} 
				>{\hsize=1.8\hsize}X  
				>{\hsize=0.8\hsize}X
				>{\hsize=1.0\hsize}X
				>{\hsize=0.7\hsize}X
				>{\hsize=0.7\hsize}X
				@{}}
			\toprule
			\textbf{Model} & \textbf{Open/Closed} & \textbf{Technique} & \multicolumn{2}{c}{\textbf{Weighted Macro F1}} \\
			\cmidrule(lr){4-5} 
			& & & \textbf{English} & \textbf{Hungarian} \\
			\midrule
			xlm-roberta-base & Open & Fine-tuning & 0.94 & 0.76 \\
			xlm-roberta-large & Open & Fine-tuning & 0.98 & 0.81 \\
			xlm-roberta-longformer-base-4096 & Open & Fine-tuning & 0.97 & 0.8 \\
			Meta-Llama-3-8B-Instruct & Open & Zero-shot & 0.64 & - \\
			gpt-3.5-turbo-0125 & Closed & Zero-shot & 0.64 & - \\
			gpt-4-turbo-2024-04-09 & Closed & Zero-shot & 0.7 & 0.49 \\
			gpt-4-turbo-2024-04-09 & Closed & One-shot & 0.72\footnotemark & - \\
			\bottomrule
		\end{tabularx}
	\end{threeparttable} \newline
	
	\footnotetext{Evaluated using 100 stratified examples due to cost constraints.}
    
	Table 2 illustrates that fine-tuned XLM-RoBERTa models significantly
	outperform zero-shot and one-shot approaches utilizing generative models
	for long text classification. Both the Longformer and the large variant
	of XLM-RoBERTa yielded comparable results, demonstrating their similar
	efficacy. Furthermore, there was no notable difference in F1 scores
	between the open Llama-3 and GPT-3.5 models. Although the one-shot
	technique with GPT-4 exhibited a slight improvement over GPT-3.5, this
	improvement may be attributed to the reduced test set size, therefore,
	might not be statistically significant.
	
	The performance gap between English and Hungarian is not particularly
	surprising. However, the fact that one-shot learning did not outperform
	zero-shot learning in this task is intriguing. One other factor
	to consider is that the dataset is taken from publicly available data,
	and it is likely that GPT models might have seen some or all of it
	during pre-training. We suggest that, despite their capability to handle
	long contexts, GPT models struggle to comprehend policy content within
	longer texts.
	
	\hypertarget{discussion}{%
		\section{Discussion}\label{discussion}}
	
	The objective of our study was to assess the impact of input text length
	on LLM performance for the multiclass classification task of the
	Comparative Agendas Project. In sum, our results suggest that, despite their capability to handle longer texts, GPT models are not competitive with BERT-based ones for low-cost research settings (which limit expenses to up to single-shot experiments with generative models). In this Discussion, we first provide suggestions for model selection for long text classification tasks based on our experimental results. Second, we dig deeper into the issue of class imbalances–a critical component for arriving at interpretable and valid results for multiclass classification applications.
	
	\hypertarget{selecting-the-right-model}{%
		\subsection{Selecting the right
			model}\label{selecting-the-right-model}}
	
	At the time the most cost-effective approach to
	employing OpenAI models for long text classification remains zero-shot
	or one-shot methods. In our experiment, these methods also clearly underperformed fine-tuned XLM-RoBERTa or Longformer models. Exploring the potential performance
	improvements with 3-shot, 5-shot, or even 10-shot approaches would be
	valuable. However, such experiments are not only prohibitively expensive
	but also challenging to execute due to the API's rate limiting. Another
	avenue for enhancing classification performance could be the fine-tuning
	of GPT specifically for the CAP classification task. This option
	requires further research and incurs similar costs to model inference.
	
	In contrast to OpenAI's proprietary models, alternatives like Llama
	demand extensive computational resources. In our fourth experiment Llama
	demonstrated performance on par with GPT-3.5 Turbo. Similar to the GPT
	models, fine-tuning Llama for specific tasks is a promising area for
	investigation. With the rapid expansion of generative model options, it
	would be valuable to evaluate very large models, such as Llama-70B or
	alternatives like Bloom or Aya, although these may require extensive
	hardware resources.
	
	Resource constraints also present a challenge with traditional encoder
	models. Despite the Longformer model scaling well in terms of memory
	usage for sequences above 512 tokens, it requires more resources than
	fine-tuning the base or even the large version of XLM-RoBERTa for
	sequences below 512 tokens. Specialized techniques, such as gradient
	accumulation, could be used to mitigate this issue. This consideration
	also suggests that adopting a specialized model like the Longformer may
	not be necessary. Instead, focusing on a more conventional model with a
	larger parameter count could be advantageous. It is also worth noting
	that comparing the classification performance of the Longformer and
	xlm-roberta-large on very long documents (longer than 4096 tokens) could
	reveal situations where the Longformer model offers a better balance
	between efficiency and performance. Table 3 summarizes the advantages
	and disadvantages of each model we tested for the CAP classification
	task.
	
	\begin{threeparttable}[H]
		\centering
		\scriptsize
        \caption{Model comparison for long text classification}
		\begin{tabularx}{\textwidth}{@{} 
				>{\hsize=0.4\hsize}X  
				>{\hsize=0.4\hsize}X  
				>{\hsize=1.1\hsize}X  
				>{\hsize=1.1\hsize}X  
				@{}}  
			\toprule
			\textbf{Model} & \textbf{Technique} & \textbf{Advantages} & \textbf{Disadvantages} \\
			\midrule
			XLM-RoBERTa\newline (base) & Fine-tuning & 
			\begin{itemize}[leftmargin=*, nosep]
				\item Fast fine-tuning and inference
				\item Fine-tuning can be done using consumer GPUs
				\item Sufficient performance
			\end{itemize} & 
			\begin{itemize}[leftmargin=*, nosep]
				\item Implementation requires some expertise
				\item Worse overall performance compared to larger variants
			\end{itemize} \\
			XLM-RoBERTa\newline (large) & Fine-tuning & 
			\begin{itemize}[leftmargin=*, nosep]
				\item Increased performance
				\item Still a small model compared to state-of-the-art generative LLMs
			\end{itemize} & 
			\begin{itemize}[leftmargin=*, nosep]
				\item Implementation requires some expertise
				\item Increased memory consumption
				\item Slower fine-tuning and inference times
			\end{itemize} \\
			Longformer (base) & Fine-tuning & 
			\begin{itemize}[leftmargin=*, nosep]
				\item Might be beneficial for very long documents (\textgreater4096 tokens)
				\item Memory-effective above 512 tokens
			\end{itemize} & 
			\begin{itemize}[leftmargin=*, nosep]
				\item Still requires special hardware
				\item Implementation needs some expertise
				\item Moderately slow inference
			\end{itemize} \\
			Llama 3 \newline(8B, Instruct) & Zero-shot & 
			\begin{itemize}[leftmargin=*, nosep]
				\item Open generative LLM
				\item Similar performance to GPT variants
			\end{itemize} & 
			\begin{itemize}[leftmargin=*, nosep]
				\item Increased memory requirements
				\item Slow inference
				\item Implementation needs some expertise
			\end{itemize} \\
			GPT-3.5 Turbo & Zero-shot & 
			\begin{itemize}[leftmargin=*, nosep]
				\item Easy access through the API
				\item Cost-effective inference
			\end{itemize} & 
			\begin{itemize}[leftmargin=*, nosep]
				\item Slow inference time
				\item Rate limiting
				\item API blackbox and downtime
			\end{itemize} \\
			GPT-4 Turbo & Zero-shot & 
			\begin{itemize}[leftmargin=*, nosep]
				\item Easy access through the API
				\item Better at complex tasks than GPT-3.5
			\end{itemize} & 
			\begin{itemize}[leftmargin=*, nosep]
				\item Slow inference time
				\item Rate limiting
				\item API blackbox and downtime
			\end{itemize} \\
			GPT-4 Turbo & One-shot & 
			\begin{itemize}[leftmargin=*, nosep]
				\item Easy access through the API
				\item The model can be provided with actual knowledge before classification
			\end{itemize} & 
			\begin{itemize}[leftmargin=*, nosep]
				\item Slow inference time
				\item Even stricter rate limiting
				\item Blackbox
				\item API downtime
				\item Very expensive due to having to show examples for each label
			\end{itemize} \\
			\bottomrule
		\end{tabularx}
	\end{threeparttable}
	
	\hypertarget{class-level-results}{%
		\subsection{Class-level results}\label{class-level-results}}
	
	For interpretable results, it is also important to explore the reasons behind misclassifications made by the models, focusing on patterns that emerge due to (i) certain categories with high support (i.e., categories with a large number of documents) and (ii) overlaps in subject matter (i.e., when policy discussions touch on multiple issues), respectively. Figure 7 displays the relevant heatmap for the class-level results of the Longformer model fine-tuned with a maximum sequence length of 2048.
	
	\customsection{High support}
	
	First, categories with a large number of documents or \emph{high
		support}, such as Government operations and Macroeconomics, tend to
	attract more misclassifications. For example, Government operations,
	which has the highest support (6,204 documents), sees substantial
	misclassifications from other categories. Specifically, 14.6 pct. of all
	documents pertaining to Civil rights are incorrectly classified as
	Government operations. Similarly, the Macroeconomic category, with a
	support of 5,010, also sees a notable normalised number of misclassifications.
	
	\customsection{Substance overlaps}
	
	Second, \emph{substance overlaps}, the substantial similarity of classes and lables in terms of their textual content, significantly contribute to these
	misclassifications (see Figure 7 for a class-level heatmap of the confusion matrix). For instance, 10.5 pct. of documents truly related
	to Labour are misclassified as Macroeconomics. This could be due to, for
	instance, discussions about labour market policies, such as wage levels
	during job training or matters related to trade unions and the
	administration of unemployment benefits. While such discussions pertain
	to the Labour issue, general discussions about the unemployment level
	and how to handle it belong to the Macroeconomics category. Thus, such
	discussions about unemployment could confuse the model and lead it to
	falsely classify documents about Labour into Macroeconomics. These errors are due to the structure of the codebook and are very difficult to mitigate strictly by the means of modeling work.
	
	Moreover, misclassifications between the Social welfare and the Labour
	issue are other cases in point (6.9 pct. of all Social welfare-related
	documents are classified as Labour, whereas 5.9 pct. of all
	Labour-related documents are falsely labelled as Social welfare). This
	could be because several policy discussions might contain aspects
	related to both issues. For instance, discussions about social security,
	poverty alleviation, or elderly care policy belong to the Social welfare
	issue, whereas \emph{work-related} transfer payments (such as early
	retirement benefits) pertain to the Labour category. Hence, policy
	debates about transfer payments, which both contain work-related and
	non-work-related payments could potentially confuse the model. In this case it is a rule of the research project ---an observation can only be assigned a single label--- that is creating difficulties for the classifier.
	
	As another example of misclassifications due to substance overlaps, 5.9
	pct. of documents truly pertaining to Civil rights were labelled as Law
	and crime. Here, some documents could contain discussions about for
	instance citizens working for foreign intelligence services or their
	protection against mass surveillance (pertaining to Civil rights) but
	also police efforts against or penalties for terrorism. That is,
	increased surveillance has been a common response to domestic terrorist
	attacks (Haggerty \& Gazso, 2005), and to the extent that these
	discussions about such responses also mention changes in penalties for
	terrorism, this could confuse the model.
	
	Finally, 4.1 pct. of Foreign trade documents experience
	misclassifications into Domestic commerce. These misclassifications
	could to some extent be due to the fact that the former issue among
	others relates to export promotion and regulation, domestic companies'
	investment in foreign countries, and regulation of imports. Such
	discussions, for instance revolving around company investment in foreign
	countries, might at the same time briefly mention conditions for small
	businesses in the country or of bank fees (both of which relate to
	Domestic commerce). Consequently, the model would falsely label the
	document as Domestic commerce rather than Foreign trade. In general,
	these misclassifications highlight the challenges faced by the model in
	differentiating between closely related policy issues. However, it
	should be noted that even for human coders, there will always be
	misclassifications due to substance overlaps, since some policy issues
	inherently overlap, and since most policy discussions contain different
	aspects of various issues.
	
	\begin{figure}[H]
		\centering
		\includegraphics[width=420pt]{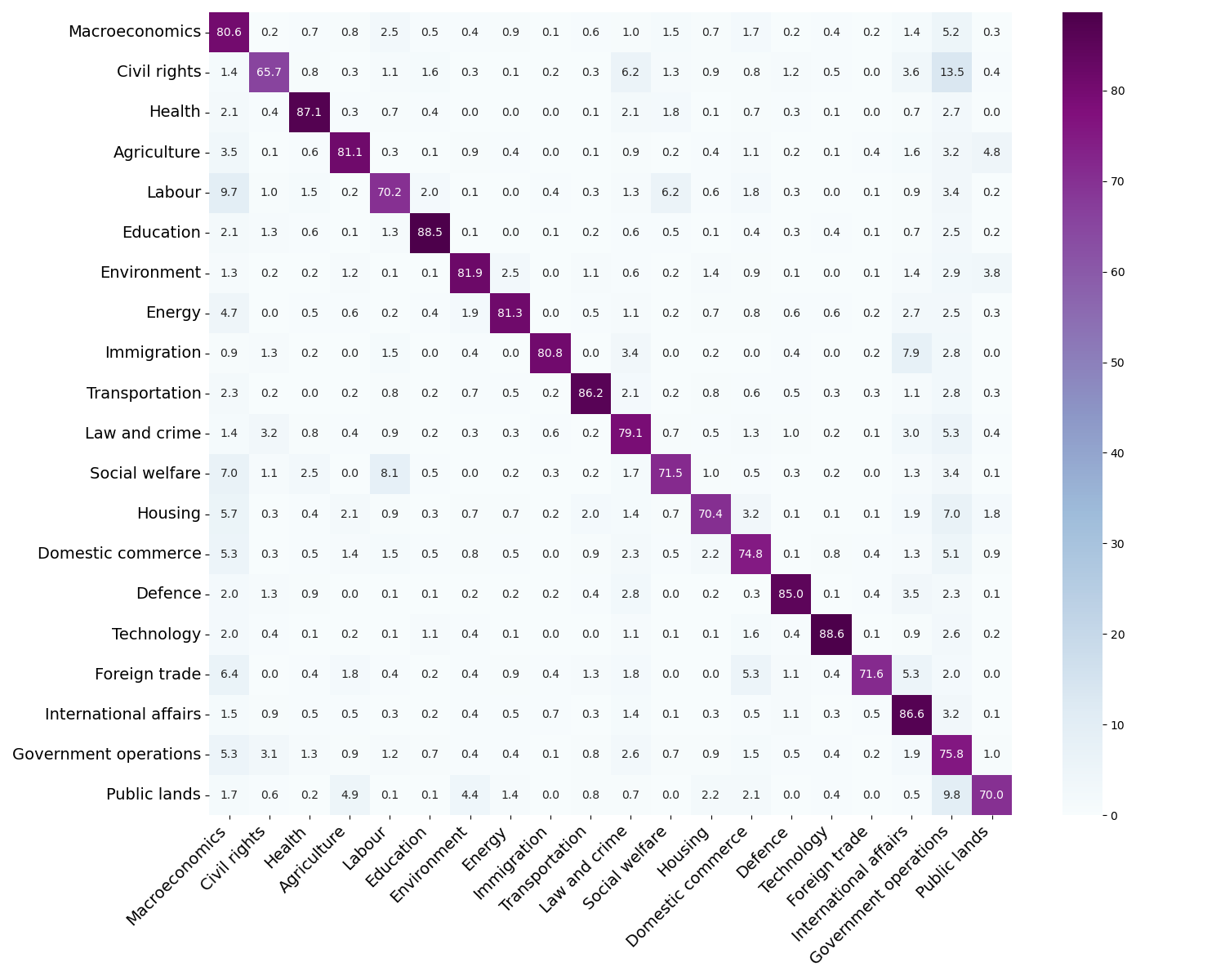}
		\caption{Heatmap of confusion matrix for the selected Longformer model}
	\end{figure}
	
	\section{Conclusion}

	In this article, we conduced experiments in relation to the classification of long, multilingual input texts according to the CAP policy topic codebook. For this task, we compared results from BET-based models with open and closed generative models. We conducted tests with GPT-3.5-Turbo and GPT-4-Turbo
	in zero and one-shot settings. While GPT-4-Turbo showed promising
	results for English text, it underperformed compared to our fine-tuned
	models, especially with Hungarian text. Conducting the full set of
	experiments using proprietary models like GPT-3.5-Turbo and GPT-4-Turbo
	would incur high costs. Although several recent open-access models, such
	as Llama \citep{touvron_llama_2023}, BLOOM \citep{bigscience_workshop_bloom_2023}, Mixtral
	\citep{jiang_mixtral_2024} and Aya \citep{ustun_aya_2024} are available, not all
	of them are pre-trained on all our languages of interest, particularly
	Hungarian. Models like Aya, which do include Hungarian, could be
	considered as cost-effective alternatives to GPT-based models though
	better performance is not guaranteed.
	
	In addition to performance, ethical considerations must be taken into account when working with generative models. While the CAP classification task is specialized, the use of generative
	LLMs like GPT and Llama raises concerns about hallucination, where the
	model generates incorrect or misleading information beyond simple
	misclassification. This issue is particularly critical when these models
	are applied in decision-making processes or tasks that influence public
	policy. To mitigate these risks, it is essential to implement robust
	validation procedures and provide uncertainty estimates alongside model
	outputs.
	
	Our study reveals that the Longformer, while outperforming the base
	XLM-RoBERTa model, did not surpass the performance of the large
	XLM-RoBERTa model despite all three models being similarly fine-tuned.
	GPT models, tested in zero-shot and one-shot settings, underperformed
	compared to the fine-tuned models. These findings suggest that for the
	CAP classification task, fine-tuned conventional models remain more
	effective than specialized or generative models in handling long text
	inputs.
	
	This study shows a way forward for scholars who wish to employ similar
	models on different documents. For instance, the content of long
	international treaties could be analyzed following our approach (see for
	example the DESTA dataset\footnotemark). Moreover, court rulings constitute another case in point. Such rulings
	could be hundreds of pages long. Consequently, scholars can benefit from
	our approach to study the content of these rulings, for instance as
	captured by the US Supreme Court dataset\footnotemark.
	
	\footnotetext{\url{https://www.designoftradeagreements.org/}}
	\footnotetext{\url{http://scdb.wustl.edu/index.php}}
	
	\newpage
	
	\customsection{Data availability statement}
	\raggedright
	Replication material is available at:\newline
	\url{https://osf.io/w3fjn/?view\_only=67372dd98f0b48349546752fee5b4e50}



	\newpage
	
	\appendix
	\section*{Appendix}
	\renewcommand{\thefigure}{A\arabic{figure}}
	\renewcommand{\thetable}{A\arabic{table}}
	\setcounter{figure}{0} 
	\setcounter{table}{0}  
	\pagenumbering{arabic} 
	
	\hypertarget{experiment-summary}{%
		\subsection*{Experiment summary}\label{experiment-summary}}
	
	Below is a detailed overview of the experiments conducted. The table
	provides a brief description of each experiment, the dependent variable,
	the fine-tuning and test data used, and the performance of each model
	(weighted macro F1). It also indicates the benchmark for each experiment
	and notes whether any model exceeded the established benchmark.
	
	{
		\vspace{2pt}
		\centering
		\scriptsize
        \captionof{table}{Summary table of the conducted experiments}
		\begin{tabularx}{\textwidth}{@{} 
				>{\hsize=0.3\hsize}X  
				>{\hsize=0.8\hsize}X
				>{\hsize=0.8\hsize}X
				>{\hsize=0.6\hsize}X
				>{\hsize=1.4\hsize}X
				>{\hsize=1\hsize}X
				@{}}
			\toprule
			\textbf{Exp.} & \textbf{Description} & \textbf{Variable} & \textbf{Options} & \textbf{Data Details} & \textbf{Results} \\
			\midrule
			E1 & Text length distribution & Fine-tuning data composition & Short & short ($<$512 tokens), 150,661 rows, 20 labels (npc dropped), stratified & Weighted Macro F1: 0.69; \newline Benchmark: short fine-tuning data; \newline Outperformed Benchmark: True \\
			\cmidrule{4-6}
			&  &  & Long & long ($\geq$512 tokens), 150,661 rows, 20 labels (npc dropped), stratified & Weighted Macro F1: 0.75 \\
			\cmidrule{4-6}
			&  &  & Mixed & mixed (1/2 short and 1/2 long), 150,661 rows, 20 labels (npc dropped), stratified & Weighted Macro F1: 0.76 \\
			\addlinespace
			E2 & Text truncation & Max sequence length & 512, 1024, 2048 & long ($\geq$512 tokens), 150,661 rows, 20 labels (npc dropped), stratified & Weighted Macro F1: 0.76 (All options) \\
			\addlinespace
			E3a & XLM-RoBERTa vs Longformer & Model type & xlm-roberta-base & mixed (1/2 short and 1/2 long), 150,661 rows, 20 labels (npc dropped), stratified & Weighted Macro F1: 0.76; \newline Benchmark: xlm-roberta-base; \newline Outperformed Benchmark: True \\
			\cmidrule{4-6}
			&  &  & longformer & long ($\geq$512 tokens), 40,000 rows, 20 labels (npc dropped), stratified & Weighted Macro F1: 0.79 \\
			\addlinespace
			E3b & Base vs Large & Model size & xlm-roberta-base, xlm-roberta-large & - & Weighted Macro F1: 0.76 (base), 0.83 (large) \\
			\addlinespace
			E4 & Proprietary vs Open & Model type & xlm-roberta-base & mixed (1/2 short and 1/2 long), 150,661 rows, 20 labels (npc dropped), stratified & Weighted Macro F1: 0.94; \newline Benchmark: 0.76; \newline Outperformed Benchmark: True \\
			\cmidrule{4-6}
			&  &  & xlm-roberta-large & mixed (1/2 short and 1/2 long), 150,661 rows, 20 labels (npc dropped), stratified & Weighted Macro F1: 0.98; \newline Benchmark: 0.81 \\	
			\bottomrule	
		\end{tabularx}
	}
	
	\newpage
	
	\hypertarget{statistical-significance}{%
		\subsection*{Statistical
			significance}\label{statistical-significance}}
	
	To determine whether the differences between model performances are
	statistically significant, we used McNemar's test to calculate P-values
	for each experiment. The test compared the models' predictions, focusing
	on cases where one model was correct and the other was incorrect. This
	process was repeated for each model combination across all experiments.
	An F1 score difference was considered statistically significant if the
	P-value was less than 0.05. Note that in E2, the differences in F1 were
	so small that it can be stated without calculating the p-value that they
	are not statistically significant.
	
	\begin{table}[H]
		\centering
		\scriptsize
        \caption{Statistical significance testing results}
		\begin{tabularx}{\textwidth}{@{} 
				>{\hsize=0.2\hsize}X  
				>{\hsize=0.4\hsize}X  
				>{\hsize=0.4\hsize}X  
				>{\hsize=0.13\hsize}X  
				>{\hsize=0.2\hsize}X  
				@{}}
			\toprule
			\textbf{Experiment} & \textbf{Model 1} & \textbf{Model 2} & \textbf{P-Value} & \textbf{Significant} \\
			\midrule
			E1 & xlm-roberta-base (short) & xlm-roberta-base (long) & 3.45e-188 & True \\
			& xlm-roberta-base (short) & xlm-roberta-base (mixed) & 7.96e-269 & True \\
			& xlm-roberta-base (mixed) & xlm-roberta-base (long) & 2.01e-05 & True \\
			E2 & longformer (512) & longformer (1024) & & False \\
			& longformer (512) & longformer (2048) & & False \\
			& longformer (1024) & longformer (2048) & &  False \\
			E3 & xlm-roberta-base & xlm-roberta-large & 7.59e-311 & True \\
			& xlm-roberta-base & longformer & 8.18e-82 & True \\
			& longformer & xlm-roberta-large & 2.60e-85 & True \\
			E4 & gpt-4-turbo-2024-04-09 (zero-shot) & longformer & 2.39e-27 & True \\
			& gpt-4-turbo-2024-04-09 \newline(zero-shot) & meta-llama-3-8b-instruct \newline(zero-shot) & 2.50e-03 & True \\
			& gpt-4-turbo-2024-04-09 \newline(zero-shot) & xlm-roberta-base & 5.07e-21 & True \\
			& gpt-4-turbo-2024-04-09 \newline(zero-shot) & xlm-roberta-large & 1.16e-27 & True \\
			& gpt-4-turbo-2024-04-09 \newline(zero-shot) & gpt-3.5-turbo-0125 \newline(zero-shot) & 5.86e-03 & True \\
			& longformer & meta-llama-3-8b-instruct \newline(zero-shot) & 3.87e-33 & True \\
			& longformer & xlm-roberta-base & 3.51e-03 & True \\
			& longformer & xlm-roberta-large & 6.28e-01 & False \\
			& longformer & gpt-3.5-turbo-0125 \newline(zero-shot) & 4.78e-33 & True \\
			& meta-llama-3-8b-instruct \newline (zero-shot) & xlm-roberta-base & 1.66e-27 & True \\
			& meta-llama-3-8b-instruct \newline (zero-shot) & xlm-roberta-large & 3.99e-34 & True \\
			& meta-llama-3-8b-instruct \newline (zero-shot) & gpt-3.5-turbo-0125 \newline(zero-shot) & 9.11e-01 & False \\
			& xlm-roberta-base & xlm-roberta-large & 2.65e-03 & True \\
			& xlm-roberta-base & gpt-3.5-turbo-0125 \newline(zero-shot) & 2.71e-26 & True \\
			& xlm-roberta-large & gpt-3.5-turbo-0125 \newline(zero-shot) & 4.84e-34 & True \\
			\bottomrule
		\end{tabularx}
	\end{table}
	
	\newpage
	
	\hypertarget{cap-labels}{%
		\subsection*{CAP labels}\label{cap-labels}}
	
	Below we provide the complete list of labels used for the CAP
	classification task, along with their respective codes and names. Please
	note that we excluded label 23 (Culture) and label 999 (No Policy
	Content) as these were not present in every language-domain combination.
	
	\begin{table}[H]
		\scriptsize
		\centering
        \caption{List of all CAP labels}
		\begin{tabular}{@{}llll@{}}
			\toprule
			\textbf{1} & Macroeconomics & \textbf{13} & Social welfare \\
			\textbf{2} & Civil Rights & \textbf{14} & Housing \\
			\textbf{3} & Health & \textbf{15} & Domestic Commerce \\
			\textbf{4} & Agriculture & \textbf{16} & Defense \\
			\textbf{5} & Labor & \textbf{17} & Technology \\
			\textbf{6} & Education & \textbf{18} & Foreign Trade \\
			\textbf{7} & Environment & \textbf{19} & International Affairs \\
			\textbf{8} & Energy & \textbf{20} & Government Operations \\
			\textbf{9} & Immigration & \textbf{21} & Public Lands \\
			\textbf{10} & Transportation & \textbf{23} & Culture (excluded) \\
			\textbf{12} & Law and Crime & \textbf{999} & No Policy Content (excluded) \\
			\bottomrule
		\end{tabular}
	\end{table}
	
	\newpage
	
	\hypertarget{data-sources}{%
		\subsection*{Data sources}\label{data-sources}}
	
	Since the data was sourced from different files for each language and
	domain, we present the distributions across source files for each
	language in Figures A1-A5. Additionally, Table A4 presents the total
	number of documents collected from the files prior to the train-test
	split.
	
	\begin{figure}[H]
		\centering
		\includegraphics[width=\textwidth]{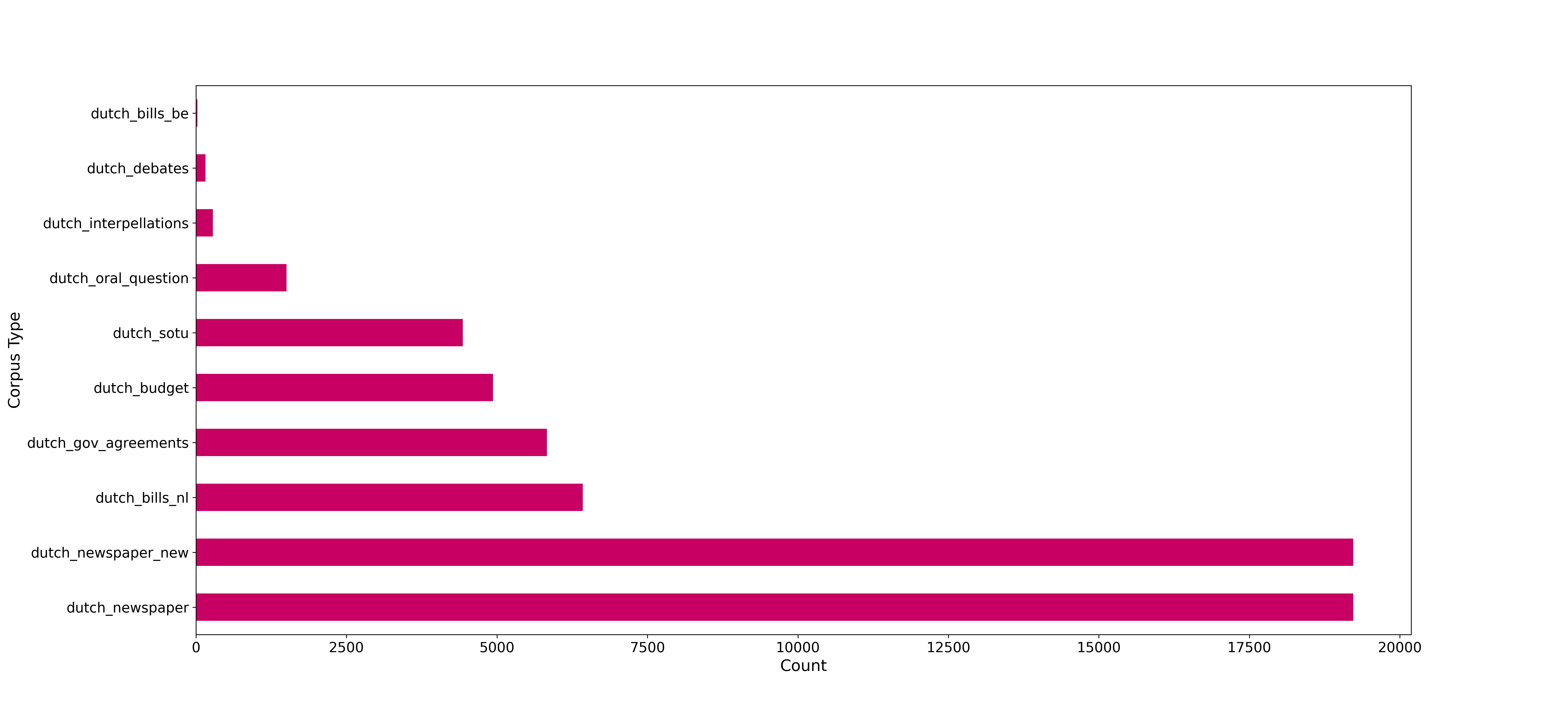}
		\caption{Amount of data by domain in Dutch}
	\end{figure}
	\begin{figure}[H]
		\centering
		\includegraphics[width=\textwidth]{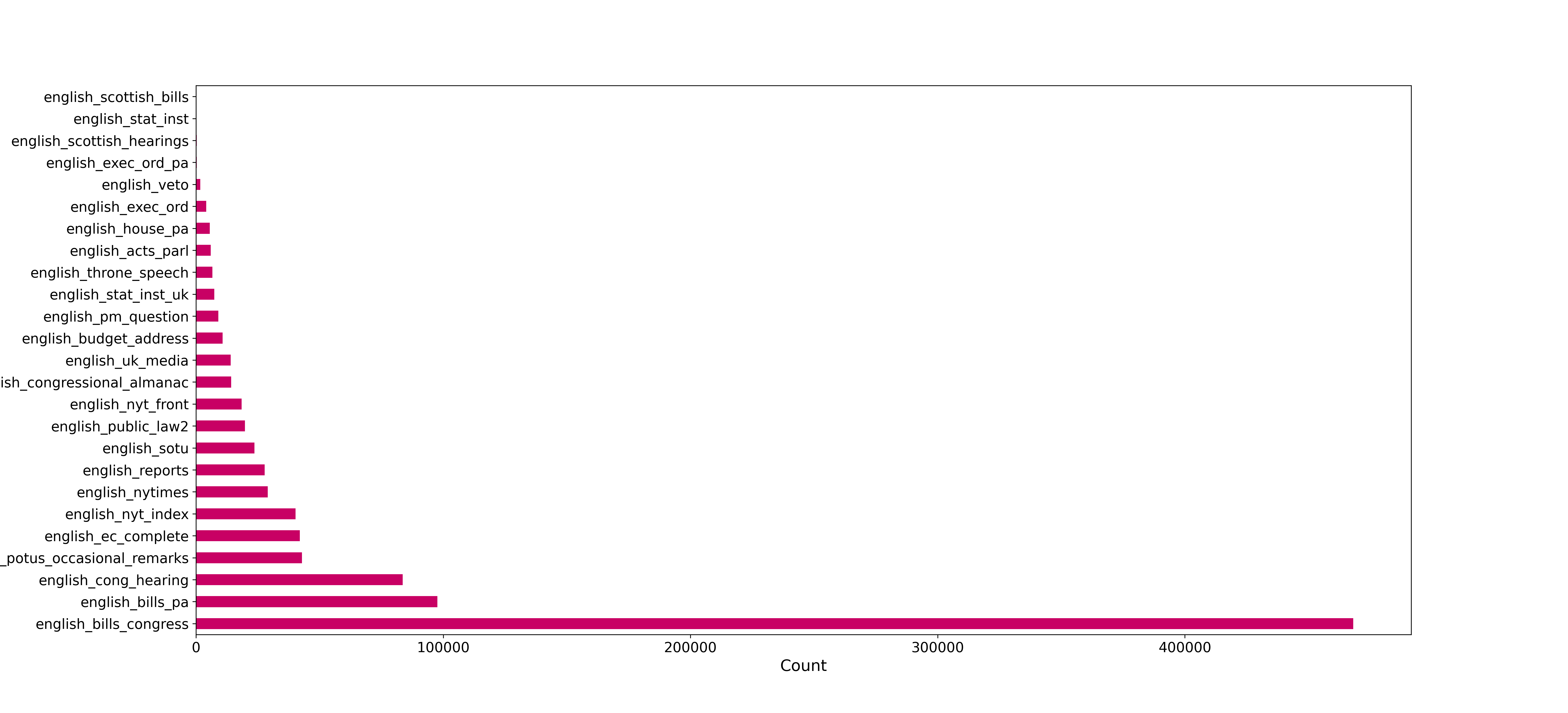}
		\caption{Amount of data by domain in English}
	\end{figure}
	\begin{figure}[H]
		\centering
		\includegraphics[width=\textwidth]{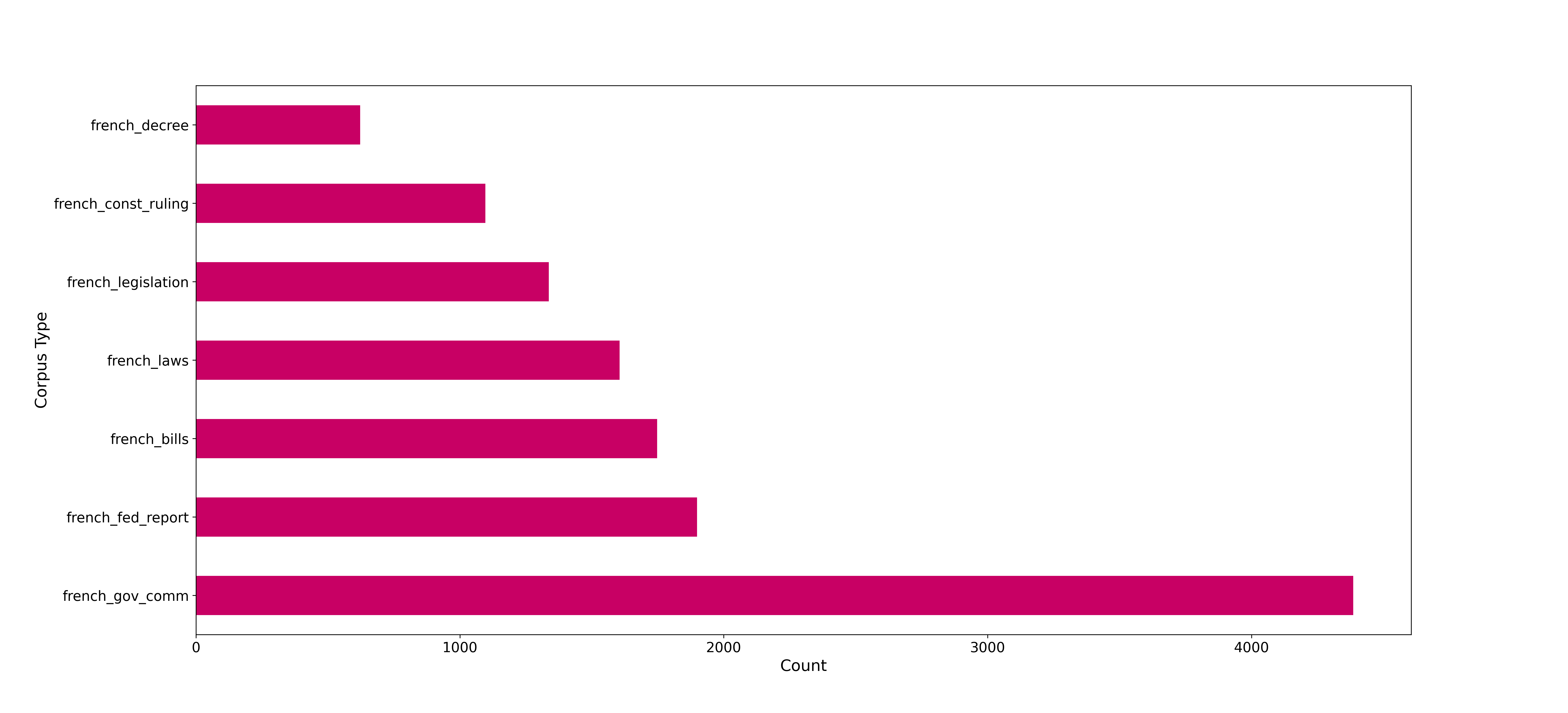}
		\caption{Amount of data by domain in French}
	\end{figure}
	\begin{figure}[H]
		\centering
		\includegraphics[width=\textwidth]{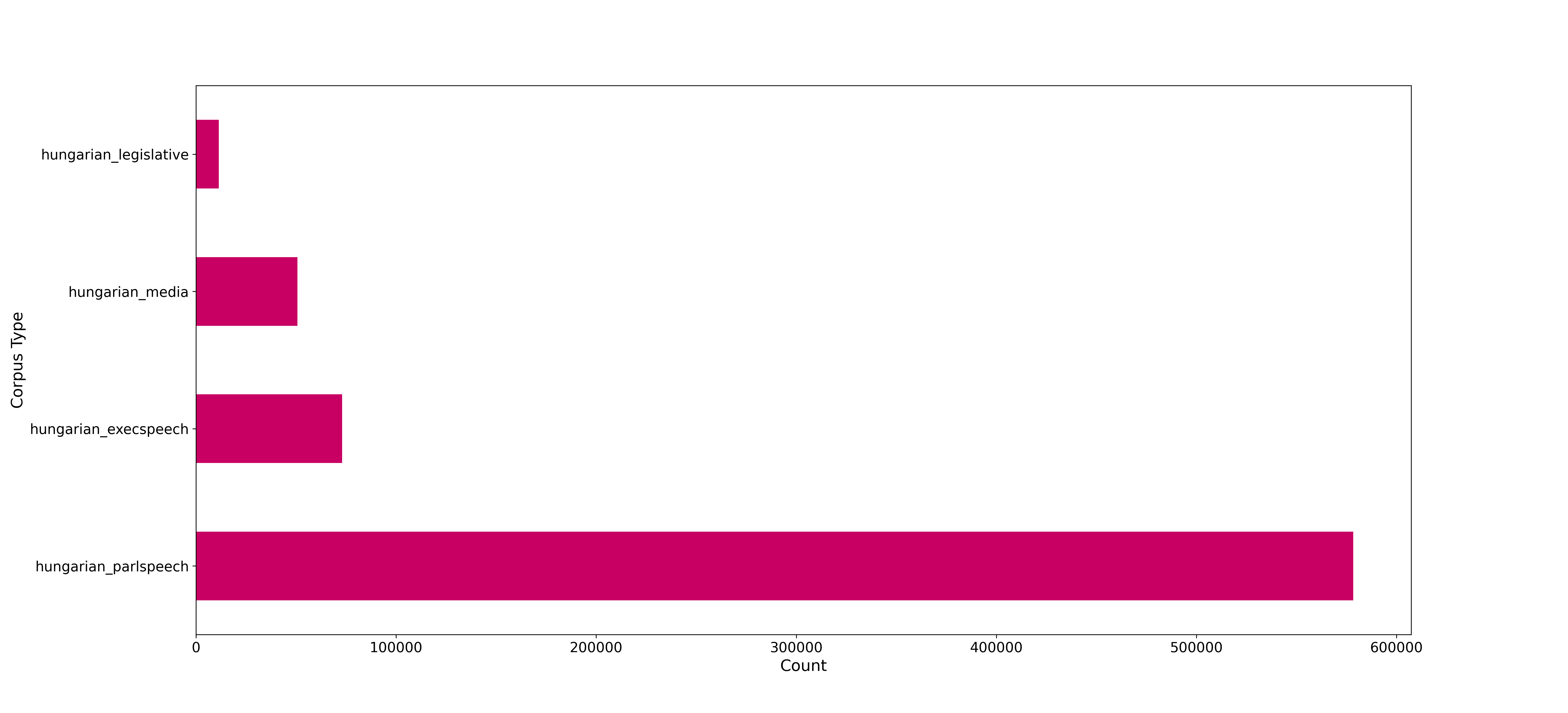}
		\caption{Amount of data by domain in Hungarian}
	\end{figure}
	\begin{figure}[H]
		\centering
		\includegraphics[width=\textwidth]{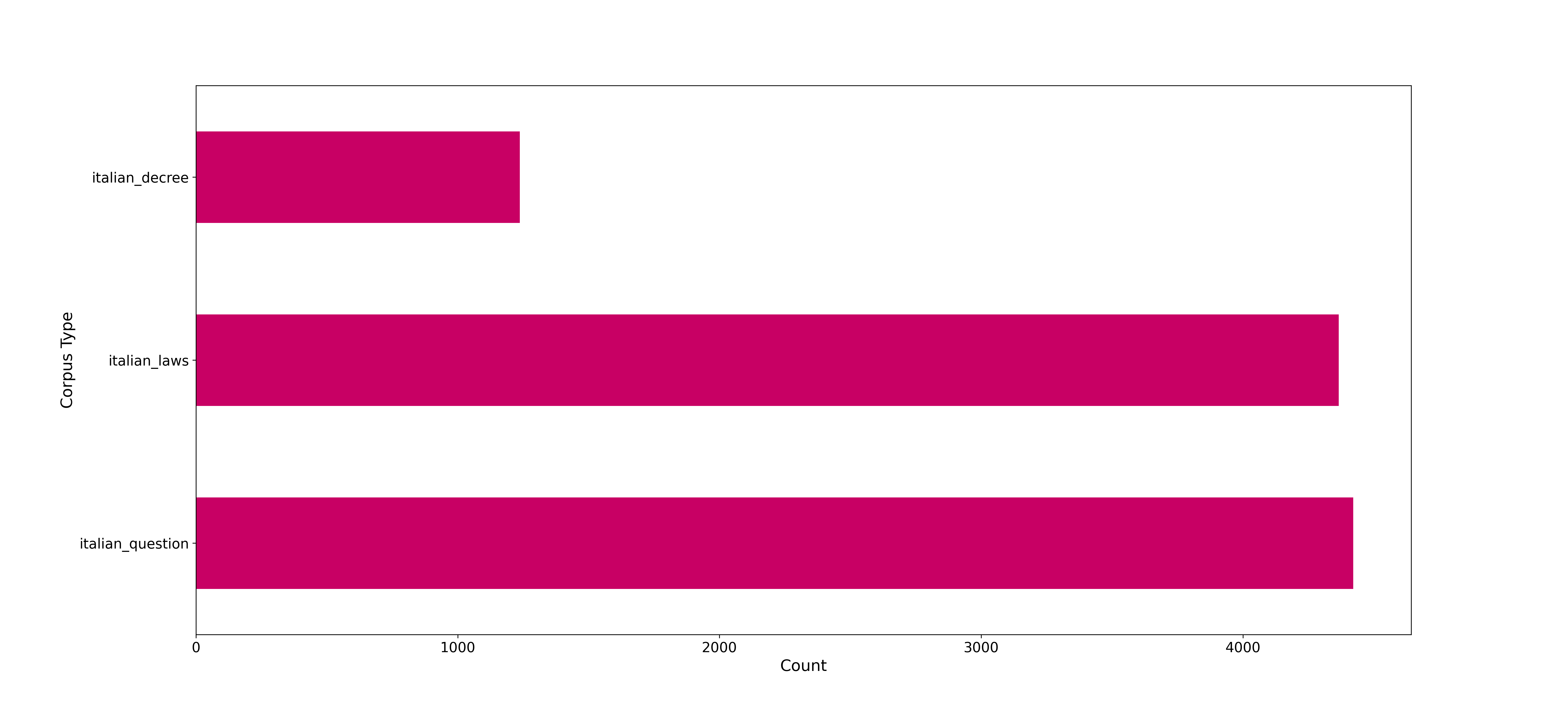}
		\caption{Amount of data by domain in Italian}
	\end{figure}
	\begin{table}[H]
		\centering
		\scriptsize
        \caption{Number of documents in the pooled dataset}
		\begin{tabular}{@{}lllll@{}}
			\toprule
			\textbf{Hungarian} & \textbf{English} & \textbf{Dutch} & \textbf{French} & \textbf{Italian} \\
			\midrule
			713,616 & 973,481 & 62,038 & 12,694 & 10,025 \\
			\bottomrule
		\end{tabular}
	\end{table}
 \label{lastpage}

\end{document}